%% file: template.tex
\documentclass[10pt,twocolumn]{article}

\usepackage[
  top=1.0in,
  bottom=1.0in,
  left=0.75in,
  right=0.75in,
  columnsep=0.25in
]{geometry}

\usepackage[utf8]{inputenc} 
\usepackage[T1]{fontenc}    
\usepackage{url}            
\usepackage{booktabs}       
\usepackage{amsfonts}       
\usepackage{nicefrac}       
\usepackage{microtype}      

\usepackage{graphicx}
\usepackage{subfig}
\usepackage{natbib}
\usepackage{doi}
\usepackage{amsmath}
\usepackage{bbding}
\usepackage{colortbl}
\usepackage{color}
\usepackage[table]{xcolor}
\usepackage{hyperref}
\usepackage{newtxtext,newtxmath}

\title{\LARGE\bfseries ARM: A Learnable, Plug-and-Play Module for CLIP-based Open-vocabulary Semantic Segmentation}

\author{
\centering
\begin{tabular}{ccc}
\textbf{Ziquan Liu} &
\textbf{Zhewei Zhu} &
\textbf{Xuyang Shi}\thanks{Corresponding author.} \\
\multicolumn{3}{c}{Southwest University of Science and Technology, China} \\
\multicolumn{3}{c}{\texttt{\{ziquanliu, zhuzhewei, xuyangshi\}@mails.swust.edu.cn}} \\
\end{tabular}
}

\begin{document}

\maketitle

\input{sec/0_abstract}    
\input{sec/1_intro}
\input{sec/2_relatedwork}
\input{sec/3_method}
\input{sec/4_experiments}
\input{sec/5_conclusion}
{
    \small
    \bibliographystyle{unsrt}
    \bibliography{references}
}

\end{document}

%% file: sec/0_abstract.tex
\begin{abstract}
Open-vocabulary semantic segmentation (OVSS) is fundamentally hampered by the coarse, image-level representations of CLIP, which lack precise pixel-level details. Existing training-free methods attempt to resolve this by either importing priors from costly external foundation models (e.g., SAM, DINO) or by applying static, hand-crafted heuristics to CLIP's internal features. These approaches are either computationally expensive or sub-optimal. We propose the Attention Refinement Module (ARM), a lightweight, learnable module that effectively unlocks and refines CLIP's internal potential. Unlike static-fusion methods, ARM learns to adaptively fuse hierarchical features. It employs a semantically-guided cross-attention block, using robust deep features (K, V) to select and refine detail-rich shallow features (Q), followed by a self-attention block. The key innovation lies in a ``train once, use anywhere" paradigm. Trained once on a general-purpose dataset (e.g., COCO-Stuff), ARM acts as a universal plug-and-play post-processor for diverse training-free frameworks. Extensive experiments show that ARM consistently boosts baseline performance on multiple benchmarks with negligible inference overhead, establishing an efficient and effective paradigm for training-free OVSS.
\end{abstract}

%% file: sec/1_intro.tex
\section{Introduction}
\label{sec:intro}

Semantic segmentation is a fundamental task in computer vision that aims to assign a precise category label to each pixel in an image. With the rise of vision-language models (VLMs), particularly CLIP \cite{clip}, this field has undergone significant transformation. In recent years, numerous studies have focused on applying CLIP to semantic segmentation tasks, leveraging its powerful zero-shot reasoning capability to generalize to categories unseen during training, thereby achieving open-vocabulary semantic segmentation.

\begin{figure}[t]
\centering
\subfloat[VFM based method]{
\includegraphics[width=0.98\linewidth]{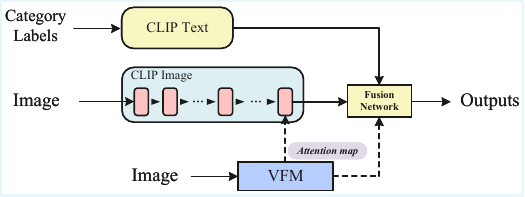}
}

\subfloat[ResCLIP]{
\includegraphics[width=0.98\linewidth]{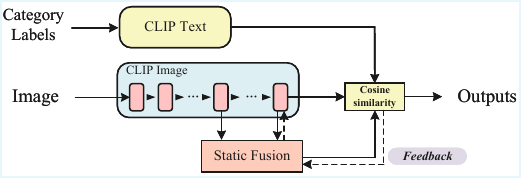}
}

\subfloat[Ours ARM-CLIP]{
\includegraphics[width=0.99\linewidth]{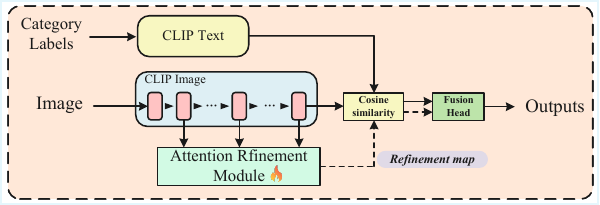}
}
\caption{Comparison of different approaches for training-free OVSS. (a) Methods relying on external Vision Foundation Models (VFMs) like SAM or DINO. (b) ResCLIP statically fuses intermediate attention maps. (c) Our ARM-CLIP learns to refine CLIP's hierarchical features using a lightweight trainable module.}
\label{fig:onecol}
\end{figure}

However, directly leveraging CLIP for segmentation faces inherent limitations \cite{maskclip, ovseg}. CLIP's pre-training objective maximizes global feature alignment, lacking explicit constraints on pixel-level local consistency. This leads to the patch-level affinity maps produced by its vision encoder being spatially coarse \cite{clip, resclip, clearclip}, which directly manifests as prevalent artifacts in segmentation results, such as blurry boundaries, loss of fine-grained details, and background noise.

To overcome this challenge, existing research has primarily evolved along two technical routes: training-based methods \cite{segclip, cat-seg, san, sed, zegclip} and training-free methods  \cite{sclip, clearclip, naclip, proxyclip, cliper, corrclip, resclip}. The former attempts to fine-tune CLIP on segmentation datasets, but this incurs substantial computational costs and risks sacrificing CLIP's inherent zero-shot generalization ability \cite{maskclip}. Consequently, training-free methods have become a research focus, endeavoring to adapt CLIP's pre-trained representations without any additional training.

Training-free methods primarily follow two paradigms. The first, as illustrated in Figure~\ref{fig:onecol}(a), relies on external Vision Foundation Models (VFMs). These methods \cite{proxyclip, cliper, corrclip, lposs, clip-dinoiser} introduce high-quality feature priors from other large-scale models \cite{sam, dino, dinov2, sd} to correct and sharpen CLIP's coarse output. Despite their effectiveness, this approach significantly increases system complexity and computational overhead.

The second, arguably more elegant paradigm focuses on mining CLIP's internal potential. This line of work can be further categorized: (1) \textit{Final-Layer Attention Modification}: These works \cite{sclip, naclip, clearclip} observed the spatial invariance of attention maps in CLIP's final layers and proposed heuristic, training-free modifications to improve localization. (2) \textit{Static fusion of internal cues}: Represented by ResCLIP \cite{resclip} as shown in Figure~\ref{fig:onecol}(b), these methods recognize that intermediate-layer attention maps retain richer spatial details. They aggregate these intermediate maps via a static, hand-crafted strategy and use them as a residual signal to correct the final coarse attention map.

Despite progress in mining CLIP's internal potential, existing strategies remain sub-optimal. Attention modification methods \cite{sclip, naclip} rely on intricate, hand-crafted heuristic designs. Concurrently, ResCLIP \cite{resclip} has shown that fusing intermediate attention maps can effectively improve segmentation performance. However, its static, hand-crafted fusion strategy is non-adaptive and underutilized the rich information available.

This leads to our core question: \textit{If a static fusion of attention maps is already beneficial, can we design a lightweight, learnable module that adaptively leverages CLIP's hierarchical features to achieve a far superior self-correction?}

Unlike training a task-specific decoder, our approach learns a universal spatial refinement without compromising CLIP’s open-vocabulary capability. To address this, we propose the Attention Refinement Module (ARM), as shown in Figure~\ref{fig:onecol}(c). The fundamental difference from ResCLIP is that ARM does not rely on static fusion of attention maps; instead, it learns to adaptively refine hierarchical features. As demonstrated in previous work \cite{dino, vit, dovit}, the ViT model constructs a rich feature hierarchy: shallow layers preserve fine-grained spatial details, such as contours and textures, while deeper layers capture abstract semantic information. Specifically, ARM extracts shallow/mid-level and deep features from the CLIP encoder. It then leverages robust deep features through a semantically guided cross-attention mechanism to filter and select detail-rich shallow features, followed by further feature fusion learning through a self-attention block. This process generates a category-independent visual refinement representation, which is then mapped to a category-specific residual refinement map via text embedding. Finally, the learned refined map is fused with the original coarse affinity map of CLIP through simple element-wise addition to obtain the final segmentation result.

We train our ARM once on the COCO-Stuff \cite{stuff} dataset and deploy it as a fixed, plug-and-play post-processor across some training-free OVSS frameworks. Extensive experiments on standard benchmarks validate our approach. 

Our main contributions are summarized as follows: 
\begin{itemize}
\item We propose the Attention Refinement Module (ARM), a lightweight, trainable module that learns to refine hierarchical features to improve CLIP's pixel-level predictions. We demonstrate its superiority over other methods, establishing a more effective path for mining VLMs internal potential.

\item We validate the generality and plug-and-play capability of ARM. Trained once, it exhibits strong zero-shot transferability, consistently boosting diverse training-free models on multiple unseen datasets.

\item Our work reveals the significant, untapped learnable potential within CLIP's internal features, offering a new, efficient paradigm for OVSS that moves beyond static heuristics and reliance on costly external foundation models. 
\end{itemize}

%% file: sec/2_relatedwork.tex
\section{Related Work}
\label{sec:related}

Research in Open-Vocabulary Semantic Segmentation (OVSS), catalyzed by the advent of CLIP \cite{clip}, has bifurcated into two principal paradigms: training-based and training-free methods. We review both, with a focus on the latter, to which our work belongs.

\noindent\textbf{Training-Based Methods.} These methods fine-tune the VLM or introduce learnable components to adapt CLIP for dense prediction tasks. Pioneering works like LSeg \cite{lseg} and OVSeg \cite{ovseg} align CLIP embeddings with pixel-level features from a segmentation backbone. ZegFormer \cite{zegformer} employs a transformer decoder for zero-shot generalization, while SAN \cite{san} introduces side adapters to efficiently fine-tune frozen backbones. ODISE \cite{odise} leverages diffusion models for panoptic mask generation, and TCL \cite{tcl} focuses on text-conditioned learning. Other notable approaches include CLIPSeg \cite{clipseg}, which adds a segmentation head to CLIP; DenseCLIP \cite{denseclip}, emphasizing dense feature alignment; SegCLIP \cite{segclip}, pre-training on synthetic data; and FC-CLIP \cite{fcclip}, fusing features across scales. GroupViT \cite{groupvit} groups visual tokens for hierarchical segmentation, and CRIS \cite{cris} uses contrastive learning for region-text alignment. While effective on some benchmarks, these methods may risk overfitting to seen classes and compromising the VLM's inherent zero-shot capabilities \cite{maskclip}, often requiring substantial computational resources for training.

\noindent\textbf{Training-Free Methods.} To preserve zero-shot generalization, these methods avoid fine-tuning the VLM, instead post-processing its outputs with auxiliary cues. Gaining traction for their efficiency, early efforts \cite{maskclip, zsseg} were the first to expose CLIP's localization limits from image-level training. Recent works mitigate this via two strategies: reliance on external foundation models or mining internal CLIP potentials.

\noindent\textbf{Training-Free Refinement via External Models.} A prevalent approach compensates for CLIP's coarse localization by importing spatial priors from external vision foundation models (VFMs) like SAM \cite{sam}, DINO/DINOv2 \cite{dino, dinov2}, or Stable Diffusion (SD) \cite{sd}. For example, PerSAM \cite{persam} personalizes SAM with CLIP prompts. CorrCLIP \cite{corrclip} uses SAM masks for scope reconstruction and DINO features for value reconstruction to suppress inter-class correlations. CLIPer \cite{cliper} employs SD for fine-grained compensation alongside internal attention fusion. FreeSeg \cite{freeseg} generates free-form masks via SD inpainting guided by CLIP. ReCo \cite{reco} uses retrieval-augmented contrast for external knowledge integration. ZegCLIP \cite{zegclip} incorporates external detectors for proposal generation. However, these methods introduce dependencies on multiple large models, increasing computational overhead and complicating inference pipelines.

\noindent\textbf{Training-Free Refinement via Internal Potential.} An alternative, more lightweight paradigm unlocks localization cues from within the frozen CLIP model, avoiding external dependencies. This line of work diagnoses issues in CLIP's Vision Transformer (ViT) \cite{vit}, where deep layers exhibit spatially-invariant attention \cite{dino, dovit}. Methods evolve along two paths.

\noindent\textit{Self-Attention Modification.} These approaches heuristically modify the final-layer attention to restore localization. GEM \cite{gem} proposes a method to compute the attention matrix by combining query-query, query-key, and value-value attention. SCLIP \cite{sclip} replaces query-key self-attention with query-query and key-key mechanisms. ClearCLIP \cite{clearclip} removes noisy residuals and feedforward neural networks in the last layer. CLIPtrase \cite{cliptrase} attempts to cluster global image patches for segmentation using a weighted average of autocorrelation attention mechanisms. NACLIP \cite{naclip} incorporates neighboring priors into key-key attention. AttCLIP \cite{attclip} enhances attention with adaptive gating. CLIP-Surgery \cite{clipsurgery} surgically corrects attention patterns by analyzing and removing noisy keys and residuals.

\noindent\textit{Fusion of Intermediate Cues.} Recognizing valuable signals in intermediate ViT layers \cite{dovit}, these methods attempt to aggregate them. ResCLIP \cite{resclip} averages intermediate maps as residuals to correct final-layer outputs. CLIPer's early-layer fusion \cite{cliper} statically averages attention maps to replace the final one. While validating the utility of intermediate cues, these methods rely on static, hand-crafted fusion strategies (e.g., simple averaging) rather than a learned, adaptive mechanism to fuse these multi-scale features.

We position ARM between static, untrained fusion and fully supervised segmentation. This makes it distinct from both static heuristics and task-specific decoders.

%% file: sec/3_method.tex
\section{Methodology}
\label{sec:Methodology}

\begin{figure*}[t]
\centering

\includegraphics[width=0.99\linewidth]{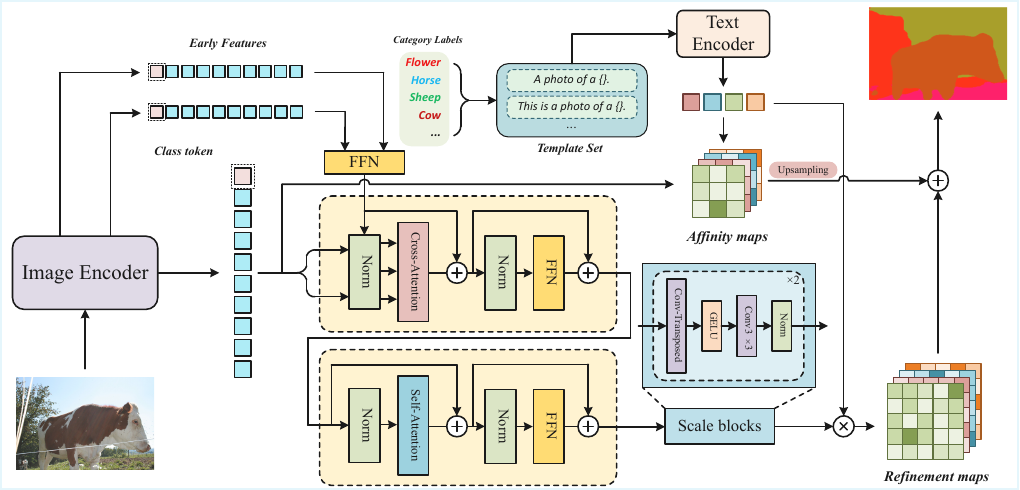}

\caption{
Our Attention Refinement Module (ARM) architecture. ARM extracts hierarchical features from the image encoder and fuses them sequentially through cross-attention and self-attention mechanisms. The fused features are processed by Scale blocks to generate a class-agnostic affinity map. This map is then multiplied by the text embedding ($\otimes$) to generate a class-related refined map. Finally, this refined map is added ($\oplus$) to the original coarse affinity map to correct localization errors and restore details.
}
\label{fig:overall}
\end{figure*}

Our core idea is to maximally exploit the features learned intrinsically by CLIP to refine its coarse segmentation, thereby avoiding the dependency on external models. We achieve this by learning a residual refinement signal derived purely from CLIP's own multi-level features. The overall pipeline, illustrated in Figure~\ref{fig:overall}, first extracts multi-level features from the CLIP visual encoder. These features are then fed into our proposed Attention Refinement Module (ARM) to generate a refinement map, which is simply fused with the original coarse affinity map in the end.

\subsection{Preliminary}

Our approach builds upon the standard training-free OVSS paradigm. Given an input image $I \in \mathbb{R}^{H \times W \times 3}$ and a set of text prompts $\mathcal{T} = \{t_1, ..., t_C\}$ for $C$ target classes.

First, we utilize the frozen CLIP text encoder to encode the prompts into text embeddings $E_{t} \in \mathbb{R}^{C \times D}$, where $D$ is the feature dimension. Simultaneously, the image $I$ is fed into the frozen CLIP ViT visual encoder. We extract the output from its final transformer block (excluding the [CLS] token) to obtain patch features $F_{d} \in \mathbb{R}^{N \times D}$, where $N = (H/P) \times (W/P)$ is the number of patches and $P$ is the patch size.

Following \cite{clip}, we generate the coarse affinity map by computing the normalized cosine similarity between $F_{d}$ and $E_{t}$, scaled by the learnable temperature coefficient $\tau$:
\begin{equation}
    S_{c} = \text{Softmax}(\tau \cdot \frac{F_d}{\|F_d\|} \cdot \frac{E_{t}^\top}{\|E_{t}\|})
\end{equation}
This affinity map $S_{c} \in \mathbb{R}^{N \times C}$ provides the basic semantic localization for the target classes, but it suffers from low spatial resolution and lacks fine-grained boundary details. We reshape and upsample it to $M_{c} \in \mathbb{R}^{C \times H' \times W'}$ to serve as the baseline for subsequent refinement.

\subsection{Attention Refinement Module}

To overcome the spatial coarseness of $M_{c}$, we designed an Attention Refinement Module (ARM), which is lightweight and transferable to other training-free methods. The core principle of this module is to perform ``self-correction" by leveraging features from different hierarchical levels within CLIP.

As \cite{cliper} and \cite{resclip} suggested that CLIP's shallow and mid-level features contain the low-level edge and texture information that is often smoothed out in the final features. We modify the forward pass of the CLIP vision encoder to extract intermediate patch features from ViT layers $l_i$ and $l_j$ (e.g., 3rd layer and 7th layer). These features are concatenated and projected through a small MLP to obtain $F_{m} \in \mathbb{R}^{N \times D}$ with a unified dimension, aligning it with $F_{d}$. The ARM takes both $F_{m}$ and $F_{d}$ as input, and its internal structure involves two key steps.

\textbf{Semantic-guided Cross-Attention.} We first use $F_{m}$ as the query and $F_{d}$ as the key and value, fusing them through a stack of cross-attention blocks. This is formulated as follows:
\begin{equation}
\begin{split}
    F'_{m} &= \text{Cross-Attn}(F_{m}, F_{d}, F_{d}) \\
&= \text{Softmax}(\frac{F_{m}\cdot F_{d}^\top}{\sqrt{d}})\cdot F_{d}
\end{split}
\end{equation}
where $d$ is the dimensionality of the keys. This step uses the robust global semantics of $F_d$ to guide and filter the low-level details in $F_m$, focusing them on semantically relevant edges and textures.

\textbf{Spatial Consistency Self-Attention.} The fused feature $F'_{m}$ is then passed through a stack of self-attention blocks to enhance internal spatial consistency and contextual awareness:
\begin{equation}
    \begin{split}
        F''_{m} &= \text{Self-Attn}(F'_{m}, F'_{m}, F'_{m}) \\
        &= \text{Softmax}(\frac{F'_{m}\cdot F'^\top_{m}}{\sqrt{d}})\cdot F'_{m}
    \end{split}
\end{equation}
After attention fusion, the resulting $F''_{m}$ is fed into a lightweight Scale Block (see Figure~\ref{fig:overall}). This module consists of transposed convolutions, which are mainly responsible for upsampling the features while restoring some spatial details of the current features. This process generates a class-independent visual representation $R_{v} \in \mathbb{R}^{D \times H' \times W'}$.

Finally, we project $R_v$ into a class-specific refinement map $M_{r} \in \mathbb{R}^{C \times H' \times W'}$ using the text embeddings $E_{t}$:
\begin{equation}
    M_{r}= E_t \cdot R_v
\end{equation}
$M_{r}$ will be the correctional signal for $M_{c}$, concentrating on boundaries and easily confused regions.

To combine the coarse global semantics from $M_{c}$ with the fine-grained local corrections from $M_{r}$ and minimize the impact on the original model, we simply fuse them using simple element-wise addition to get an initial fused map $M_{f}$:
\begin{equation}
    M_{f} = M_{c} + M_{r}
\end{equation}
This residual connection allows the refinement module to focus on learning the correction signal rather than reconstructing the entire segmentation map from scratch. 




%% file: sec/4_experiments.tex
\section{Experiments}
\label{sec:Experiments}

\subsection{Experimental Setups}
\label{sec:setup}
\noindent\textbf{Datasets.} We train our ARM once on the COCO-Stuff (Stuff) \cite{stuff}. For zero-shot evaluation, we use five standard benchmarks: PASCAL VOC 2012 (VOC-20/21) \cite{voc}, PASCAL Context (Context-59/60) \cite{context}, ADE20K (ADE-150) \cite{ade}. And in Section~\ref{sec:ablation_study}, we also used the COCO-Object (Object) \cite{stuff} dataset for ablation. We compare all the semantic segmentation results using mean Intersection over Union (mIoU).

\noindent\textbf{Architecture.} Our experiments primarily use the CLIP ViT-B/16 backbone, with additional results provided for ViT-L/14. The CLIP model itself remains frozen during both ARM training and inference. By default, our ARM extracts intermediate features from 3rd layer and 7th layer, comprising one cross-attention block and one self-attention block.

\noindent\textbf{Training details.} We train only the ARM using the Binary Cross-Entropy (BCE) loss on COCO-Stuff \cite{stuff} for 5 epochs. We use the AdamW \cite{adamw} optimizer with a learning rate of $1 \times 10^{-4}$, a batch size of 4, and a weight decay of $1 \times 10^{-3}$. 

\noindent\textbf{Inference.} During evaluation, the trained ARM is frozen and applied as a post-processor to the baseline affinity map.

\subsection{Main Results}
\label{sec:transfer}

\begin{figure}[h!]
\centering
\includegraphics[width=0.95\linewidth]{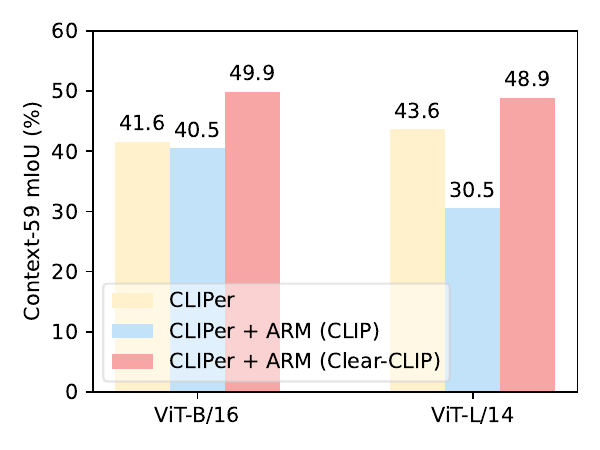}
\caption{Zero-shot transfer results of applying ARM to CLIPer on the Context-59. The ARM trained with standard CLIP features (blue) resulted in a performance degradation, while the ARM trained with Clear-CLIP features (red) resulted in a significant performance improvement.}
\label{fig:clip_clearclip}
\end{figure}

Initially, we trained the ARM using the original CLIP \cite{clip}, and the results were acceptable. However, when transferring it to other training-free methods, performance dropped noticeably. We noticed that most current training-free methods use a Clear-CLIP \cite{clearclip} configuration, such as removing the residual connections and feedforward neural networks in the last layer. Therefore, we first trained the ARM under the Clear-CLIP configuration and then transferred it to CLIPer, achieving better performance, as shown in Figure~\ref{fig:clip_clearclip}.

\begin{table*}[h!]
\centering
\caption{Reference Comparison with Training-Based Methods. We provide this comparison to contextualize the performance of our lightweight, 'train-once' module against fully-trained methods.}
\label{tab:vs_training_based}
\resizebox{0.95\linewidth}{!}
{
\begin{tabular}{l|c|c|cccccc}
\toprule
Method & Backbone & Training Set & Context-459 & Context-59 & VOC-20 & ADE-150 & VOC-21 \\
\midrule
TCL \cite{tcl}  & ViT-B/16 & CC-12M & - & 33.9 & 83.2 & 17.1 & 55.0 \\
OVSeg \cite{ovseg} & ViT-B/16 & COCO-Stuff & 11.0 & 53.3 & 92.6 & 24.8 & - \\
CAT-Seg \cite{cat-seg} & ViT-B/16 & COCO-Stuff & 19.0 & 57.5 & 94.6 & 31.8 & 77.3 \\
SAN \cite{san} & ViT-B/16 & COCO-Stuff & 12.6 & 53.8 & 94.0 & 27.5 & - \\
SED \cite{sed} & ConvNeXt-B & COCO-Stuff & 18.6 & 57.3 & 94.4 & 31.6 & - \\
ESC-Net \cite{esc-net} & ViT-B/16 & COCO-Stuff & 21.1 & 59.0 & 97.3 & 35.6 & 80.1 \\

\rowcolor{gray!20}
ARM-CLIP (Ours) & ViT-B/16 & COCO-Stuff & 9.6 & 49.9 & 89.3 & 17.8 & 73.3  \\
\midrule
OVSeg \cite{ovseg} & ViT-L/14 & COCO-Stuff & 11.0 & 53.3 & 92.6 & 24.8 & - \\
CAT-Seg \cite{cat-seg} & ViT-L/14 & COCO-Stuff & 19.0 & 57.5 & 94.6 & 31.8 & 82.5 \\
SAN \cite{san} & ViT-L/14 & COCO-Stuff & 12.6 & 53.8 & 94.0 & 27.5 & - \\
SED \cite{sed} & ConvNeXt-L & COCO-Stuff & 18.6 & 57.3 & 94.4 & 31.6 & - \\
ESC-Net \cite{esc-net} & ViT-L/14 & COCO-Stuff & 21.1 & 59.0 & 97.3 & 35.6 & 86.3 \\

\rowcolor{gray!20}
ARM-CLIP (Ours) & ViT-L/14 & COCO-Stuff & 10.6 & 50.9 & 92.6 & 18.7 & 69.3 \\
\bottomrule
\end{tabular}
}
\end{table*}

\begin{table*}[h!]
\centering
\caption{Performance comparison with training-free state-of-the-art methods. We apply the trained ARM to training-free methods using zero-shot transfer. Note that the ARM is trained using the Clear-CLIP setup.}
\label{tab:transfer}
\resizebox{0.95\linewidth}{!}{
\begin{tabular}{l|c|ccccc|c}
\toprule
Method & Backbone & Context-60 & Context-59 & VOC-20 & ADE-150 & VOC-21 & Average \\
\midrule
CLIP \cite{clip} & ViT-B/16 & 8.4 & 9.2 & 41.9 & 2.9 & 16.4 & 15.8
\\ 
MaskCLIP \cite{maskclip} & ViT-B/16 & 23.6 & 26.4 & 74.9 & 9.8 & 38.8 & 34.7
\\
ClearCLIP \cite{clearclip} & ViT-B/16 & 32.6 & 35.9 & 80.9 & 17.7 & 51.8 & 43.8
\\
NACLIP \cite{naclip} & ViT-B/16 & 35.0 & 38.4& 83.0& 19.1& 64.1& 47.9
\\
CorrCLIP \cite{corrclip} & ViT-B/16 & 44.2& \underline{48.8}& \underline{88.8}& \textbf{26.9}& \textbf{74.8}& \textbf{56.7}
\\
ResCLIP \cite{resclip} & ViT-B/16 & 33.5& 36.8& 86.0& 18.0& 61.1& 47.0
\\
SCLIP \cite{sclip} & ViT-B/16 & 34.2 & 34.2 & 80.4 & 16.1 & 59.1 & 44.8
\\
CLIPer \cite{cliper} & ViT-B/16 & 37.6& 41.7& 85.2& \underline{21.4}& 65.9& 50.4
\\
\rowcolor{gray!20}
\textbf{SCLIP + ARM} & ViT-B/16 & \underline{45.3} & 48.1 & 88.5 & 17.1 & 69.5 & 53.7 \textcolor[HTML]{00B800}{(+8.9)}
\\
\rowcolor{gray!20}
\textbf{CLIPer + ARM} & ViT-B/16 & \textbf{45.6} & \textbf{49.5} & \textbf{89.3} & 17.5 & \underline{73.2} & \underline{55.0} \textcolor[HTML]{00B800}{(+4.6)}
\\
\midrule
CLIP \cite{clip} & ViT-L/14 & 4.1 & 4.4 & 15.6 & 1.7 & 8.2 & 6.8
\\ 
MaskCLIP \cite{maskclip} & ViT-L/14 & 11.7 & 12.4 & 29.4 & 7.2 & 23.3 & 16.8
\\
ClearCLIP \cite{clearclip} & ViT-L/14 & 26.7 & 29.6 & 80.0 & 15.0 & 46.1 & 39.5
\\
CorrCLIP \cite{corrclip} & ViT-L/14 & \underline{44.9}& \textbf{50.8}& \underline{91.5}& \textbf{30.7}& \textbf{76.7}& \textbf{58.9}
\\
ResCLIP \cite{resclip} & ViT-L/14 & 30.9& 34.5& 85.5& 18.2& 54.1& 44.6
\\
SCLIP \cite{sclip} & ViT-L/14 & 22.3 & 25.2 & 69.1 & 10.9 & 43.5 & 34.2
\\
CLIPer \cite{cliper} & ViT-L/14 & 38.0& 43.6& 90.0& \underline{24.4}& 69.8& 53.2
\\
\rowcolor{gray!20}
\textbf{SCLIP + ARM} & ViT-L/14 & 42.9 & 46.9 & 89.8 & 17.5 & 65.8 & 52.6 \textcolor[HTML]{00B800}{(+18.4)}
\\
\rowcolor{gray!20}
\textbf{CLIPer + ARM} & ViT-L/14 & \textbf{46.3} & \underline{50.7} & \textbf{92.5} & 18.2 & \underline{71.6} & \underline{55.9} \textcolor[HTML]{00B800}{(+2.7)}
\\
\bottomrule
\end{tabular}
}
\end{table*}

\noindent\textbf{Reference comparison with training-based methods.} To position our method within the broader OVSS landscape, we first present a reference comparison in Table~\ref{tab:vs_training_based} between our ARM (applied to the ClearCLIP baseline) and current mainstream training-based methods.

We first wish to clarify a significant paradigm disparity: These training-based methods are fully end-to-end trained architectures that are deeply optimized for the segmentation task. In contrast, our ARM is a lightweight plug-and-play module designed with a greater emphasis on maintaining generality and efficiency. This difference in training paradigms is predictably reflected in performance. As expected, in dense scenarios like ADE-150 and Context-459, the training-based models exhibit a significant performance ceiling. We believe this gap is reasonable due to the complexity and specialization of the models.

\noindent\textbf{Quantitative results of ARM in zero-shot transfer.} Table~\ref{tab:transfer} illustrates the zero-shot transfer capability of ARM as a plug-and-play module. Our ARM delivers significant mIoU gains for all baseline methods. For the ViT-B/16 architecture, ARM improves the average mIoU of SCLIP from 44.8\% to 53.7\% and CLIPer from 50.4\% to 55.0\%. This improvement is equally robust on the ViT-L/14 architecture. This demonstrates the potential of ARM as a general-purpose post-processing module.

Additionally, the introduction of ARM results in a slight performance degradation of the model on the ADE-150 dataset. We attribute this to the inherent complexity of the ADE-150 dataset, which places exceptionally high demands on a model's feature extraction capabilities. In contrast, our ARM is designed with a lightweight and concise architecture. This compact design may limit its capacity to learn the rich features required for fine-grained segmentation in such complex scenarios. Consequently, the refinement map generated by ARM may conflict with or interfere with the segmentation results of baseline models, leading to a slight performance degradation. Despite the limitations mentioned above, our method achieved a competitive level in the remaining benchmarks.

\begin{figure*}[h!]
\centering
\includegraphics[width=0.95\linewidth]{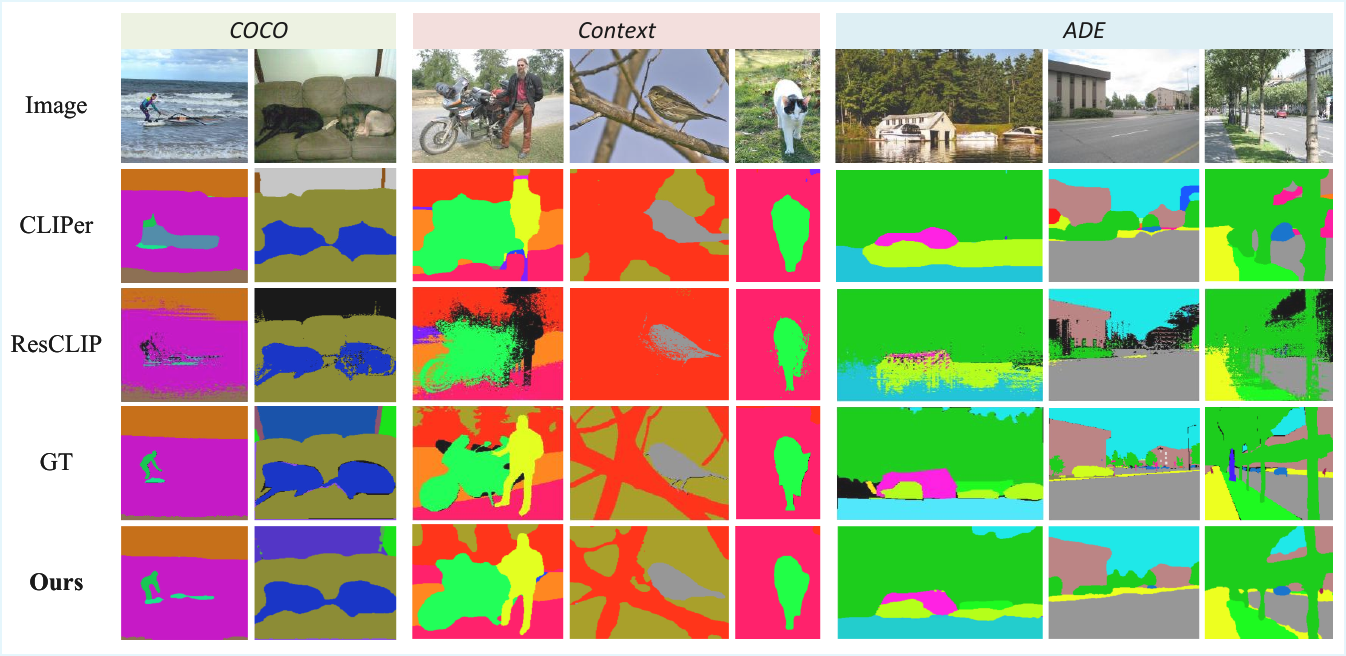}
\caption{\textbf{Qualitative Results.} Visual comparison on sample images from three different datasets. Compared to these methods, our method has more accurate segmentation results that are closer to the ground-truths.}
\label{fig:qualitative}
\end{figure*}

\noindent\textbf{Qualitative results.} As shown in Figure~\ref{fig:qualitative}, these examples demonstrate that our method offers some visual improvements. For instance, while CLIPer's output captures the general global semantics, it has limitations in capturing object boundaries and fine textures. When zero-shot transfer is applied to our ARM, the baseline model's performance in these areas is significantly improved; boundary regions appear clearer, with reduced background noise. In contrast, while ResCLIP also attempts to utilize intermediate features, its static fusion strategy and the subsequent PAMR \cite{pamr} post-processing appear to introduce artifacts or lose details in certain situations.

This suggests that ARM, as a general, learnable refinement module, with its adaptive feature extraction paradigm, may offer a valuable supplement to improving the detail performance of baseline models compared to static fusion strategies, demonstrating its effectiveness and potential.

\begin{table}[h!]
\centering
\caption{Efficiency in reporting the average inference time per image on a single NVIDIA Tesla V100 GPU.}
\label{tab:efficiency}
\resizebox{0.95\linewidth}{!}
{
\begin{tabular}{l|c|c|c}
\toprule
Method & Backbone & Time (ms) & Context-59 \\
\midrule
NACLIP & ViT-B/16 & 84.2 & 35.2 \\
ResCLIP & ViT-B/16 & 64.5 & 36.8 \\
CLIPer & ViT-B/16 & 35.4 & 41.7 \\
\rowcolor{gray!20}
\textbf{Ours} & ViT-B/16 & \textbf{53.2} \textcolor{red}{(+20.8)} & \textbf{49.5} \textcolor[HTML]{00B800}{(+8.8)} \\

\midrule
NACLIP & ViT-L/14 & 101.9 & 32.1 \\
ResCLIP & ViT-L/14 & 80.8 & 34.5 \\
CLIPer & ViT-L/14 & 67.8 & 43.6 \\
\rowcolor{gray!20}
\textbf{Ours} & ViT-L/14 & \textbf{79.7} \textcolor{red}{(+11.9)} & \textbf{50.8} \textcolor[HTML]{00B800}{(+7.2)} \\
\bottomrule
\end{tabular}
}
\end{table}

\noindent\textbf{Efficiency Analysis.}
Table~\ref{tab:efficiency} shows that our ARM adds minimal computational overhead while delivering substantial mIoU improvements. This positions our method as a highly efficient alternative to external model fusion and a more performant option compared to static internal fusion methods like ResCLIP.

\begin{table}[h!]
\centering
\caption{Ablation on intermediate layer selection.}
\label{tab:layers_ablation}
\resizebox{0.95\linewidth}{!}
{
\begin{tabular}{c|ccc}
\toprule
Fusion Layers & Context-59 & ADE-150 & Object \\
\midrule
(1st,\quad 3rd) &  45.2& 15.2 & 55.3 \\
(1st,\quad 5th) & 47.9 & 16.8 & 59.4 \\
(1st,\quad 7th) & 48.2 & 17.0 & 61.8 \\
(3rd,\quad 5th) & 47.9 & 16.5 & 59.5 \\
(\textbf{3rd,\quad 7th}) & \textbf{49.5} & \textbf{17.5} & \textbf{63.5} \\
(5th,\quad 7th) & 48.6 & 16.9 & 62.6 \\
\bottomrule
\end{tabular}
}
\end{table}

\subsection{Ablation Studies}
\label{sec:ablation_study}
All ablation experiments were performed by transferring our trained ARM to the CLIPer architecture, and were evaluated on Context-59 using ViT-B/16 as the backbone, unless otherwise specified. Our goal was to validate key design choices in the ARM.

\noindent\textbf{Choice of Intermediate Layers.}
We first investigate the impact of fusing features from different depths of the ViT backbone. Generally, fusing features from different levels usually yields better results, and our data also demonstrates this, as shown in Table~\ref{tab:layers_ablation}. For example, using the feature maps of the 1st layer and 3rd layer results in a lack of semantic features, ultimately achieving mIoU of 45.2\%, 15.2\%, and 55.3\% on the Context-59, ADE-150, and Object datasets, respectively. However, Combining a shallow layer, which may preserve more spatial details, with a deep layer containing richer semantic context yields optimal performance. As the experimental results show, using features from the 3rd layer and 7th layer yields better results.

\begin{table}[h!]
\centering
\caption{Ablation on the number of self-attention and cross-attention layers.}
\label{tab:ablation_layers}
\resizebox{0.95\linewidth}{!}{
\begin{tabular}{c|cc}
\toprule
Number of attention layers & Context-59 & Object \\
\midrule
(0,\quad 0) & 42.2 & 51.8 \\
(0,\quad 1) & 43.1 & 53.5 \\
(1,\quad 0) & 45.7 & 55.2 \\
(\textbf{1,\quad 1}) & \textbf{49.5} & \textbf{63.5} \\
(2,\quad 1) & 49.1 & 62.1 \\
(1,\quad 2) & 49.5 & 62.3 \\
(2,\quad 2) & 49.4 & 63.2 \\
\bottomrule
\end{tabular}
}
\end{table}

\begin{table}[h!]
\centering
\caption{Ablation on both attention layers and Scale Blocks.}
\label{tab:ablation_scale}
\resizebox{0.95\linewidth}{!}{
\begin{tabular}{c|c|c}
\toprule
Number of attention layers & Scale blocks & Context-59 \\
\midrule
(0,\quad 0) & \XSolidBrush & 40.1 \\
(0,\quad 0) & \XSolidBrush & 42.2 \\
(1,\quad 1) & \XSolidBrush & 46.6 \\
\rowcolor{gray!20}
(\textbf{1,\quad 1}) & \Checkmark & \textbf{49.5} \\ 
\bottomrule
\end{tabular}
}
\end{table}

\noindent\textbf{ARM Component Analysis.} Table~\ref{tab:ablation_layers} shows the contribution of self-attention and cross-attention in ARM. ($x$, $y$) means there are $x$ self-attention layers and $y$ cross-attention layers. We observed that stacking more layers did not yield further gains and slightly reduced performance. As shown in Table~\ref{tab:ablation_layers}, increasing the number of layers from (1, 1) to (2, 1), (1, 2), or (2, 2) resulted in almost no change in the mIoU score, or only a very slight decrease. This indicates that simply increasing the module depth does not provide additional performance gains. Given that increasing the number of layers incurs additional parameters and computational costs without a corresponding performance return, we therefore selected (1, 1) as the optimal configuration, which aligns with our goal to design efficient and lightweight modules.

We also validate the contribution of Scale Blocks in Table~\ref{tab:ablation_scale}. Removing both the attention module and the scale blocks results in a baseline of 40.1\% mIoU. Adding just the scale blocks provides a modest improvement by 2.1\% mIoU. Using only the attention module without scale blocks brings a substantial gain to 46.6\% mIoU, demonstrating it is the core of the refinement. Finally, combining both the attention module and the scale blocks achieves our best performance. This suggests that while the attention mechanism performs the primary feature refinement, the learnable upsampling blocks are also beneficial for generating the final high-quality refinement map.

%% file: sec/5_conclusion.tex
\section{Conclusion}
\label{sec:conclusion}
In this paper, we propose ARM, a lightweight, learnable, plug-and-play module. ARM learns how to adaptively fuse hierarchical features within CLIP, leveraging the robust semantics of deep features to guide and filter detail-rich shallow features. The core innovation lies in the ``train once, use everywhere" paradigm. ARM, trained once on a benchmark dataset, can serve as a general-purpose post-processor, enabling zero-shot transfer to various training-independent frameworks. More importantly, the inference overhead resulting from this performance improvement is negligible. Our work highlights the learnable potential inherent in CLIP’s internal features, and explores a practical paradigm for lightweight, training-free open-vocabulary segmentation.